\documentclass[conference]{IEEEtran}
\IEEEoverridecommandlockouts
\usepackage{cite}
\usepackage{amsmath,amssymb,amsfonts}
\usepackage{algorithmic}
\usepackage{graphicx}
\usepackage{textcomp}
\usepackage{xcolor}
\usepackage{hyperref}
\def\BibTeX{{\rm B\kern-.05em{\sc i\kern-.025em b}\kern-.08em
    T\kern-.1667em\lower.7ex\hbox{E}\kern-.125emX}}

\usepackage{xcolor}

\newcommand{\huggingfacesmall}{\includegraphics[width=9px]{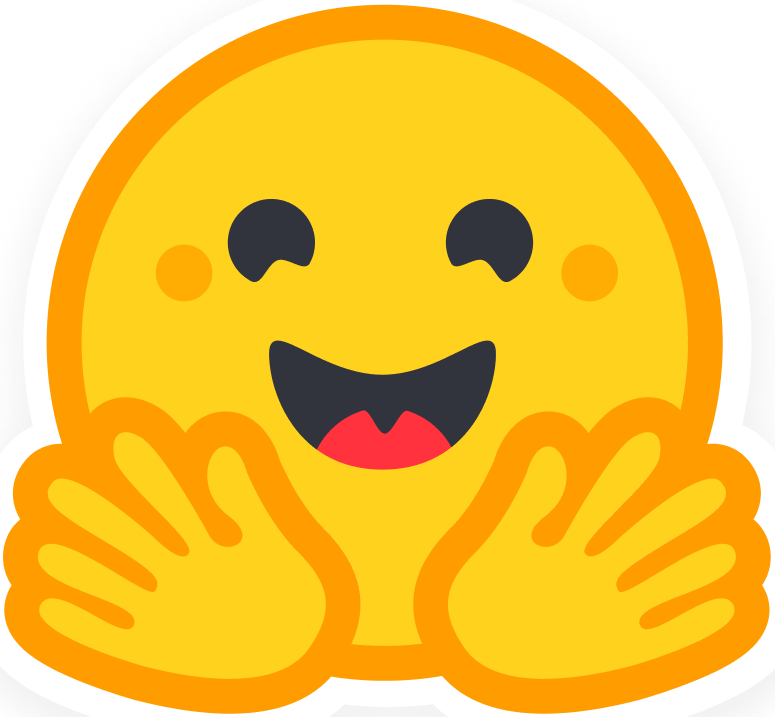}}

\usepackage{booktabs}

\begin{document}
\title{Speech-to-SOAP: End-to-End Summarization of Medical Dialogues:
KIT@BeTraC 2026}
%\title{Conference Paper Title*\\
%{\footnotesize \textsuperscript{*}Note: Sub-titles are not captured in Xplore and
%should not be used}
%\thanks{Identify applicable funding agency here. If none, delete this.}
%}

\author{\IEEEauthorblockN{1\textsuperscript{st} Enes Yavuz Ugan}
\IEEEauthorblockA{\textit{ISL, Karlsruhe Institute of Technology} \\
%\textit{Karlsruhe Institute of Technology}\\
%Karlsruhe, Germany \\
enes.ugan@kit.edu}
\and
\IEEEauthorblockN{2\textsuperscript{nd} Fabian Retkowski}
\IEEEauthorblockA{\textit{ISL, Karlsruhe Institute of Technology} \\
%\textit{Karlsruhe Institute of Technology}\\
%City, Country \\
%email address or ORCID
}
\and
\IEEEauthorblockN{3\textsuperscript{rd} Yuka Ko}
\IEEEauthorblockA{\textit{AI4LT, Karlsruhe Institute of Technology} \\
%\textit{Karlsruhe Institute of Technology}\\
%City, Country \\
%email address or ORCID
}
\and
\IEEEauthorblockN{4\textsuperscript{th} Thai-Binh Nguyen}
\IEEEauthorblockA{\textit{ISL, Karlsruhe Institute of Technology} \\
%\textit{Karlsruhe Institute of Technology}\\
%City, Country \\
%email address or ORCID
}
\and
\IEEEauthorblockN{5\textsuperscript{th} 
Maike Züfle}
\IEEEauthorblockA{\textit{AI4LT, Karlsruhe Institute of Technology} \\
%\textit{Karlsruhe Institute of Technology}\\
%City, Country \\
%email address or ORCID
}
\and
\IEEEauthorblockN{6\textsuperscript{th} Jan Niehues}
\IEEEauthorblockA{\textit{AI4LT, Karlsruhe Institute of Technology} \\
%\textit{Karlsruhe Institute of Technology}\\
%City, Country \\
%email address or ORCID
}
\and
\IEEEauthorblockN{7\textsuperscript{th} Alexander Waibel}
\IEEEauthorblockA{\textit{InterACT, Carnegie Mellon University} \\
\textit{ISL, Karlsruhe Institute of Technology}\\
%email address or ORCID
}
}

\maketitle

\begin{abstract}
With the advent of Large Language Models and its instruction following capabilities a promising application is the task of summarization.
Within this domain of task the extractive sub-task of clinical protocolling has emerged as a topic of particular interest as it can significantly reduce the downtime and protocolling burden of health-care workers thus enabling them to focus on their core work helping humans.
%Even further automating this task is directly protocolling from speech input without the use of intermediate transcriptors, further reducing downtime and opening up new avenues for future protocolling systems enabling the protocolling of meta information such as coughing which might be overlooked when using dialogue transcripts.
A further step towards automation is the direct generation of clinical notes from speech without intermediate transcripts, reducing processing time while preserving information such as coughing or other paralinguistic cues that may be lost in transcript-based systems.
To this end, we present KIT's submission to this years BeTraC challenge in the lightweight track.
Our main contribution is a scalable data augmentation pipeline that unifies heterogeneous medical dialogue datasets through synthetic speech generation and automatically generated SOAP supervision, enabling robust adaptation of a speech foundation model for end-to-end speech-to-SOAP generation.
\end{abstract}


%\section{Introduction}
%Some intro
\vspace{-5pt}
\section{Experimental Setup}
\vspace{-5pt}
%\subsection{Architecture}
\paragraph{Architecture}
We use Qwen2.5-Omni-3B~\cite{xu2025qwen25omnitechnicalreport}, an end-to-end speech-language model that has demonstrated instruction following~\cite{papi2025mcif} and speech summarization~\cite{retkowski2025summarizing} capabilities. The model is adapted to medical dialogue understanding and SOAP note generation using LoRA \cite{pham2021efficient, hu2022lora} in LLaMA-Factory~\cite{zheng2024llamafactory}.
\vspace{-1pt}
%\subsection{Data and Data Augmentation}
\paragraph{Data and Data Augmentation}
%To improve the robustness and generalization of our medical speech-language model, we combine several publicly available medical dialogue and SOAP note datasets. Since these datasets differ considerably in modality, annotation granularity, and target outputs, we convert all resources into a unified instruction-following format compatible with the Qwen2.5-Omni chat template.
%Tables~\ref{tab:data_statistics} summarize the characteristics of all datasets used in this work.

%We use Synth-DoPaCo \cite{labrak2026generating}, ACI-Bench \cite{aci-bench}, MTS-Dialog \cite{mts-dialog}, PriMock57 \cite{korfiatis2022primock57}, and OMI \cite{wang2024notechat} (Table~\ref{tab:data_statistics}).
We use Synth-DoPaCo \cite{labrak2026generating}, ACI-Bench \cite{aci-bench}, MTS-Dialog \cite{mts-dialog}, PriMock57 \cite{korfiatis2022primock57}, and OMI \cite{wang2024notechat} datasets. 
Synth-DoPaCo and OMI are fully synthetic datasets, ACI-Bench contains role-played encounters, PriMock57 consists of simulated consultations with real recordings, and MTS-Dialog provides text-only medical conversations. For datasets without audio, we synthesize speech using Kokoro-82M~\cite{li2023styletts}. All datasets are converted into a unified Audio→SOAP, Transcript→SOAP, and Audio→Diarized Transcript format when applicable.
The resulting data consists of 18795 dialogues and a total of 1653,067 hours of audio.
The final dataset and code are publicly available. \footnote{\href{https://github.com/enesyugan/IWSLTFactory}{enesyugan/IWSLTFactory}}
\footnote{\huggingfacesmall{}\href{https://huggingface.co/datasets/YapayNet/betrac2026-augmented}{YapayNet/betrac2026-augmented}}

% [ADD] main SOAP generation
% For datasets without SOAP-style target notes, we generate additional SOAP supervision from dialogue transcripts using Qwen-3.5-27B in a non-thinking setting. Each prompt consists of a general task instruction and optional SOAP formatting guidance. For downstream training, we use a SOAP template + concept statistics prompt, which combines an annotated SOAP structure and representative examples with corpus-derived clinical concept frequencies, charting-style expressions, and lay-to-clinical terminology guidance. This setting aims to normalize heterogeneous note styles and terminology into a consistent SOAP format. 
For datasets without SOAP-style target notes, we generate additional SOAP supervision from dialogue transcripts using the open-weight Qwen3.5-27B\footnote{\url{https://huggingface.co/Qwen/Qwen3.5-27B}} model served locally with vLLM in a non-thinking setting, with a maximum context length of 32,768 tokens. For downstream training, we use a SOAP template + concept statistics prompt combining an annotated SOAP structure and examples with corpus-derived clinical concept frequencies, charting-style expressions, and lay-to-clinical terminology guidance. This aims to normalize heterogeneous note styles and terminology into a consistent SOAP format.
\vspace{-5pt}
\subsection{Experiments}
We conduct several experiments to understand the impact of pretraining, data modality, and reasoning strategies on medical SOAP note generation.
All results are collated in Tables~\ref{tab:prompt_ablation}-\ref{tab:final_systems}.

\paragraph{Prompts}
We tried 3 different system prompts in order to determine their effect on the base models performance and to choose which one to use.
Additionally we applied two of the more complex prompts in the actual user instruction of the model.

\paragraph{Duration and Cleaning.}
During manual inspection, we observed several severe TTS hallucinations where the generated audio no longer matched the reference transcript. We therefore developed an alignment-based script that removes non-aligned audio segments.
To study the effect of long conversations, we additionally filtered training samples longer than 15, 21, and 25 minutes in the DoPaCo dataset, motivated by the reported average dialogue duration of 9 minutes.
%\paragraph{Duration and Cleaning.}
%Listening to the audio we heard several cases of severe TTS hallucination, thus we developed a script that aligns audio with transcripts and cuts out hallucinated parts of the audio, not aligned with the transcript.
%Additionally, to determine if longer audios are usufull or hurt performance as the average duration was reported to be around 9mins we did initial experimants on the DoPaCo data with excluded that are longer then 15min, 21min and 25min.

\paragraph{Multi-stage Methods.}
We investigate different initialization and multi-stage adaptation strategies for adapting Qwen2.5-Omni to medical dialogue understanding. In particular, we study whether intermediate tasks such as Transcript$\rightarrow$SOAP and Audio$\rightarrow$ASR improve subsequent Audio$\rightarrow$SOAP generation.
We additionally evaluate whether explicit speaker diarization benefits downstream SOAP generation. However, after Audio$\rightarrow$ASR adaptation, the model already achieves a speaker-attributed WER of approximately 3\%, indicating strong inherent diarization capabilities. Consequently, we do not pursue explicit speaker diarization in subsequent experiments.

\paragraph{Training on Speech versus Text and Speech.}
A central question of this work is whether transcript supervision improves end-to-end speech-to-SOAP generation. We therefore compare models trained exclusively on Audio$\rightarrow$SOAP examples with models jointly trained on Audio$\rightarrow$SOAP and Transcript$\rightarrow$SOAP data.
Joint training may provide additional supervision by decoupling speech recognition from the downstream summarization task while exposing the model to a larger number of semantic SOAP generation examples.

\paragraph{Chain-of-Thought Generation.}
We investigate chain-of-thought (CoT) supervision by introducing intermediate reasoning targets before SOAP note generation. Specifically, we derive several reasoning targets tailored to the evaluation metrics, including medical concepts, entities, and terminology extraction. We further compare explicit reasoning traces enclosed in \texttt{<think>}...\texttt{</think>} tags with natural-language reasoning prompts preceding the final SOAP note generation.

%Medical note generation often requires reasoning over symptoms, diagnoses, medications, and temporal information scattered throughout the conversation. To encourage more structured reasoning, we investigate chain-of-thought (CoT) generation strategies.

%In particular, we explore prompting strategies in which the model first predicts intermediate medical concepts, entities, or terminology before generating the final SOAP note. The resulting notes are then compared against direct end-to-end SOAP generation without explicit reasoning.

%\todo{\textcolor{blue}{Fabian} Describe the exact CoT prompting format and whether reasoning traces are retained or removed during inference.}

%\todo{\textcolor{blue}{Fabian} Describe all investigated CoT variants and provide examples in the appendix.}

\subsection{Training Details}
Unless otherwise stated, all experiments employ the official Qwen2-Omni chat template, FlashAttention~2, bfloat16 precision, and gradient checkpointing. We apply LoRA with rank $r=32$ to all target modules while keeping the multimodal projector frozen throughout training.

Models are optimized using AdamW with a learning rate of $1\times10^{-4}$, cosine learning rate decay, and a warmup ratio of $10\%$. Following \cite{ugan2026multilingual}, we use a small effective batch size of $4$, which also allows training under our computational constraints. Model selection is performed using the checkpoint with the lowest development-set perplexity.
%All experiments are conducted using parameter-efficient fine-tuning with LoRA rank $r=32$. We employ the official Qwen2-Omni chat template, FlashAttention~2, bfloat16 precision, and gradient checkpointing throughout all experiments.

%Unless otherwise specified, models are trained using the AdamW optimizer with a learning rate of $1\times10^{-4}$, cosine learning rate decay, and a warmup ratio of $10\%$.

%In \cite{ugan2026multilingual} it was shown that using a small batchsize is enough, due to limited ressources and time constraints we decided for an effective batchsize of 4.
%For evaluation we selected the checkpoint with lowest perplexity during training.

\subsection{Evaluation}

We follow the official BeTraC evaluation protocol for medical SOAP note generation.
The primary evaluation metrics are based on lexical overlap measures, including ROUGE, which quantify the similarity between generated and reference SOAP notes. However, lexical metrics alone may not fully capture clinical correctness, as semantically equivalent notes can differ substantially in wording.

\section{Results}
\paragraph{Prompt Ablation.}
%\begin{table}[t]
%\centering
%\scriptsize
%\setlength{\tabcolsep}{4pt}
%\renewcommand{\arraystretch}{1.1}
%\begin{tabular}{lccc}
%\toprule
%\textbf{System} & \textbf{Concept-F1} & \textbf{R-2} & \textbf{R-3} \\
%\midrule
%\multicolumn{4}{l}{\textit{Baselines}} \\
%Reported & 0.2604 & 0.0920 & 0.0344 \\
%Base Prompt & \textbf{0.3276} & 0.1315 & 0.0589 \\
%\midrule
%\multicolumn{4}{l}{\textit{System Prompt Ablation}} \\
%Sys-D & 0.2348 & 0.1232 & 0.0610 \\
%Sys-DE & 0.2332 & 0.1171 & 0.0569 \\
%\midrule
%\multicolumn{4}{l}{\textit{Instruction Prompt Ablation}} \\
%Ins-D & 0.2666 & \textbf{0.1671} & \textbf{0.0919} \\
%Ins-DE & 0.2521 & 0.1482 & 0.0780 \\
%\bottomrule
%\end{tabular}
%\caption{Development-set prompt ablations. ``D'' denotes a detailed prompt and ``DE'' a detailed prompt with an example. ``Sys'' and ``Ins'' indicate system and instruction prompting, respectively.}
%\label{tab:prompt_ablation}
%\end{table}
We investigate the impact of prompt design and prompt placement on SOAP note generation. Starting from a simple baseline prompt, we evaluate more detailed prompts both as system prompts (\texttt{System}) and as instruction prompts (\texttt{Instruction}), with and without additional examples.
%\vspace{-4pt}
\begin{table}[!h]%[htbp]
\caption{Development-set prompt ablation. Detailed prompts were evaluated as system or instruction prompts, with and without an example.}
\scriptsize
\begin{center}
\begin{tabular}{|l|c|c|c|}
\hline
\textbf{System} & \textbf{Concept-F1} & \textbf{R-2} & \textbf{R-3} \\
\hline
\multicolumn{4}{|l|}{\textit{Baselines}} \\
\hline
Reported & 0.2604 & 0.0920 & 0.0344 \\
\hline
Base Prompt & \textbf{0.3276} & 0.1315 & 0.0589 \\
\hline
\multicolumn{4}{|l|}{\textit{System Prompt Ablation}} \\
\hline
System-Detailed & 0.2348 & 0.1232 & 0.0610 \\
\hline
System-Detailed+Example & 0.2332 & 0.1171 & 0.0569 \\
\hline
\multicolumn{4}{|l|}{\textit{Instruction Prompt Ablation}} \\
\hline
Instruction-Detailed & 0.2666 & \textbf{0.1671} & \textbf{0.0919} \\
\hline
Instruction-Detailed+Example & 0.2521 & 0.1482 & 0.0780 \\
\hline
\end{tabular}
\label{tab:prompt_ablation}
\end{center}
\end{table}
As shown in Table~\ref{tab:prompt_ablation}, increasing prompt complexity in the system prompt consistently degrades performance compared to the simple baseline. In contrast, placing the detailed prompts to the instruction position substantially improves summarization quality, achieving the best ROUGE-2 and ROUGE-3 scores. These results suggest that Qwen2.5-Omni is sensitive to instruction placement and benefits more from detailed task guidance as explicit instructions rather than as persistent system-level behavior.
\paragraph{Training on Speech versus Text and Speech.}

We investigate whether incorporating transcript-conditioned supervision improves downstream SOAP note generation. Table~\ref{tab:pretraining_ablation} compares models trained solely on Audio$\rightarrow$SOAP examples with models jointly trained on Audio$\rightarrow$SOAP and Transcript$\rightarrow$SOAP data.

Joint audio-text training consistently improves clinical concept extraction, increasing Concept-F1 from 0.4780 to 0.4902, while yielding nearly identical ROUGE scores. This suggests that transcript supervision primarily benefits the model's ability to identify and represent medically relevant information rather than substantially changing the lexical overlap with reference notes. Consequently, we adopt joint Audio+Text$\rightarrow$SOAP training in our subsequent experiments.

\paragraph{Multi-Stage Adaptation.}

We further investigate whether intermediate adaptation tasks can improve downstream Audio$\rightarrow$SOAP generation. In particular, we consider intermediate objectives including Audio$\rightarrow$ASR, Transcript$\rightarrow$SOAP, joint Audio/Text$\rightarrow$SOAP training, and chain-of-thought (CoT) supervision.

As shown in Table~\ref{tab:pretraining_ablation}, all intermediate adaptation strategies substantially outperform the Audio$\rightarrow$ASR baseline. Initializing from an Audio$\rightarrow$ASR model and subsequently fine-tuning on Audio$\rightarrow$SOAP achieves the strongest ROUGE-2 and ROUGE-3 scores, indicating that explicit transcript generation provides a useful intermediate representation for medical note generation. In contrast, CoT supervision achieves the highest Concept-F1 score, suggesting that explicitly modeling intermediate reasoning steps can improve the extraction of clinically relevant concepts. In general, no single adaptation strategy dominates across all metrics, but intermediate adaptation consistently provides large improvements over direct Audio$\rightarrow$ASR training.
\begin{table}[!h]
\caption{Development-set ablations for multi-stage adaptation, joint audio/text training, and duration/audio cleaning. A, T, and AT denote audio, transcript, and audio/transcript inputs, respectively. "Clean" indicates transcript-aligned audio after removing hallucinated TTS segments.}
\begin{center}
\scriptsize
\begin{tabular}{|c|l|c|c|c|}
\hline
\textbf{\#} & \textbf{Strategy / Dataset} & \textbf{C-F1} & \textbf{R-2} & \textbf{R-3} \\
\hline

\multicolumn{5}{|l|}{\textit{Audio $\rightarrow$ SOAP}} \\
\hline
1 & A-SOAP & 0.4780 & 0.3366 & 0.2283 \\
\hline

\multicolumn{5}{|l|}{\textit{Audio + Text $\rightarrow$ SOAP}} \\
\hline
2 & AT-SOAP & 0.4902 & 0.3366 & 0.2261 \\
\hline

\multicolumn{5}{|l|}{\textit{Multi-Stage}} \\
\hline
3 & A-ASR & 0.3233 & 0.0940 & 0.0407 \\
\hline
4 & T-SOAP $\rightarrow$ A-SOAP & 0.4834 & 0.3261 & 0.2188 \\
\hline
5 & A-ASR $\rightarrow$ A-SOAP & 0.4871 & \textbf{0.3430} & \textbf{0.2338} \\
\hline
6 & A-ASR $\rightarrow$ AT-SOAP & 0.4906 & 0.3378 & 0.2275 \\
\hline
7 & A-ASR + AT-SOAP & 0.4842 & 0.3256 & 0.2175 \\
\hline
8 & AT-SOAP $\rightarrow$ A-SOAP & 0.4809 & 0.3234 & 0.2151 \\
\hline
9 & AT-CoT $\rightarrow$ AT-SOAP & \textbf{0.4908} & 0.3391 & 0.2275 \\
\hline

\multicolumn{5}{|l|}{\textit{Duration and Audio Cleaning}} \\
\hline
10 & Clean (21 min) & 0.4898 & 0.3368 & 0.2278 \\
\hline
11 & Clean (25 min) & 0.4791 & 0.3319 & 0.2218 \\
\hline
12 & Unclean (15 min) & 0.4781 & 0.3287 & 0.2210 \\
\hline
13 & Unclean (21 min) & \textbf{0.4965} & 0.3386 & 0.2287 \\
\hline
14 & Unclean (25 min) & 0.4780 & 0.3366 & 0.2283 \\
\hline

\end{tabular}
\label{tab:pretraining_ablation}
\end{center}
\end{table}

\paragraph{Duration and Cleaning Ablation.}
During manual inspection of the synthetic DoPaCo audio, we observed several severe TTS hallucinations in which the generated audio diverged substantially from the reference transcript. To mitigate this issue, we developed an alignment-based cleaning procedure that removes audio segments that cannot be aligned with the corresponding transcript.

We further investigate the effect of long conversations by filtering training examples exceeding 15, 21, and 25 minutes in duration. As shown in Table~\ref{tab:pretraining_ablation}, cleaning does not improve performance, with the best results obtained using the uncleaned dataset filtered to a maximum duration of 21 minutes. Increasing the duration threshold beyond 21 minutes yields no further improvements and slightly degrades performance, suggesting that very long conversations introduce additional noise that outweighs the benefit of increased training data.
%\begin{table}[htbp]
%\caption{Duration and audio-cleaning ablation on the development set. ``Clean'' denotes datasets after removing hallucinated TTS segments using transcript-audio alignment.}
%\begin{center}
%\scriptsize
%\begin{tabular}{|c|l|c|c|c|}
%\hline
%\textbf{\#} & \textbf{Dataset} & \textbf{Concept-F1} & \textbf{R-2} & \textbf{R-3} \\
%\hline
%10 & Clean (21 min)   & 0.4898 & 0.3368 & 0.2278 \\
%\hline
%11 & Clean (25 min)   & 0.4791 & 0.3319 & 0.2218 \\
%\hline
%12 & Unclean (15 min) & 0.4781 & 0.3287 & 0.2210 \\
%\hline
%13 & Unclean (21 min) & \textbf{0.4965} & \textbf{0.3386} & \textbf{0.2287} \\
%\hline
%14 & Unclean (25 min) & 0.4780 & 0.3366 & 0.2283 \\
%\hline
%\end{tabular}
%\label{tab:duration_ablation}
%\end{center}
%\end{table}
%\vspace{-5pt}
%\begin{table}[t]
%\centering
%\scriptsize
%\setlength{\tabcolsep}{4pt}
%\renewcommand{\arraystretch}{1.1}
%\begin{tabular}{clccc}
%\toprule
%\textbf{\#} & \textbf{Dataset} & \textbf{Concept-F1} & \textbf{R-2} & \textbf{R-3} \\
%\midrule
%10 & Clean (21 min)   & 0.4898 & 0.3368 & 0.2278 \\
%11 & Clean (25 min)   & 0.4791 & 0.3319 & 0.2218 \\
%12 & Unclean (15 min) & 0.4781 & 0.3287 & 0.2210 \\
%13 & Unclean (21 min) & \textbf{0.4965} & \textbf{0.3386} & \textbf{0.2287} \\
%14 & Unclean (25 min) & 0.4780 & 0.3366 & 0.2283 \\
%\bottomrule
%\end{tabular}
%\caption{Duration and audio-cleaning ablation on the development set. ``Clean'' denotes datasets after removing hallucinated TTS segments using transcript-audio alignment.}
%\label{tab:duration_ablation}
%\end{table}
\paragraph{Chain-of-Thought Generation.}
We investigated several chain-of-thought (CoT) strategies based on intermediate medical concepts, entities, terminology extraction, and explicit reasoning traces. Although natural-language reasoning performed better than explicit \texttt{<think>} tags, none of the explored CoT variants improved over direct end-to-end SOAP generation. Consequently, all final systems use direct SOAP generation.
%We investigated several chain-of-thought (CoT) prompting strategies, including intermediate medical concept generation, explicit \texttt{<think>}...\texttt{</think>} reasoning traces, and natural-language reasoning prompts preceding SOAP generation. Although natural-language reasoning performed better than explicit thinking tags, none of the explored CoT strategies improved upon direct end-to-end SOAP generation. Consequently, all final systems use direct SOAP generation without explicit reasoning supervision.

\paragraph{Final Systems.}
After selecting the most promising training strategies through the preceding ablations, we train our final systems using all available datasets and generated supervision. Table~\ref{tab:final_systems} reports two representative models together with our final submission.

Motivated by \cite{izmailov2018averaging,ugan2025weight,vander2023rehearsal}, we further investigate checkpoint averaging using different combinations of the best-performing development-set checkpoints. Among the evaluated combinations, averaging the checkpoints corresponding to rows \texttt{13}, \texttt{16}, and \texttt{17} achieved the best development-set performance and was therefore selected for our final submission.
%Based on \cite{izmailov2018averaging, ugan2025weight, vander2023rehearsal} we additionally experimented with averaging several combinations of the best-performing checkpoints on the development set. Among the investigated combinations, the checkpoints corresponding to rows \texttt{13}, \texttt{16}, and \texttt{17} yielded the strongest development performance and were therefore selected for our final submission.
%The resulting merged model consistently outperforms all individual systems, achieving a Concept-F1 of 0.4986, an R-2 score of 0.3537, and an R-3 score of 0.2417.
\vspace{-5pt}
\begin{table}[!h]%[htbp]
\caption{Representative final systems, merged development-set submission model, and official submission results.}
\begin{center}
\scriptsize
\begin{tabular}{|c|l|c|c|c|}
\hline
\textbf{\#} & \textbf{System / Split} & \textbf{C-F1} & \textbf{R-2} & \textbf{R-3} \\
\hline
\multicolumn{5}{|l|}{\textit{Development-set model selection}} \\
\hline
15 & A-ASR $\rightarrow$ AT-SOAP & 0.4918 & 0.3345 & 0.2256 \\
\hline
16 & AT-CoT $\rightarrow$ AT-SOAP & 0.4894 & 0.3403 & 0.2300 \\
\hline
17 & AT-SOAP & 0.4924 & 0.3430 & 0.2307 \\
\hline
18 & Merged Submission & \textbf{0.4986} & \textbf{0.3537} & \textbf{0.2417} \\
\hline

\multicolumn{5}{|l|}{\textit{Official primary submission (Row 18)}} \\
\hline
-- & DoPaCo test & 0.4949 & 0.3601 & 0.2499 \\
\hline
-- & Mock dialogue & 0.4618 & 0.3186 & 0.2011 \\
\hline
-- & Realistic & 0.4855 & 0.3430 & 0.2326 \\
\hline

\multicolumn{5}{|l|}{\textit{Official contrastive submission (Row 13)}} \\
\hline
-- & DoPaCo test & 0.3889 & 0.2161 & 0.1289 \\
\hline
-- & Mock dialogue & 0.3372 & 0.1771 & 0.0940 \\
\hline
-- & Realistic & 0.2814 & 0.1377 & 0.0733 \\
\hline

\end{tabular}
\label{tab:final_systems}
\end{center}
\end{table}
%\begin{table}[htbp]
%\caption{Representative final systems and the merged submission model on the development set.}
%\begin{center}
%\scriptsize
%\begin{tabular}{|c|l|c|c|c|}
%\hline
%\textbf{\#} & \textbf{System} & \textbf{C-F1} & \textbf{R-2} & \textbf{R-3} \\
%\hline
%10 & A-ASR $\rightarrow$ AT-SOAP & 0.4918 & 0.3345 & 0.2256 \\
%\hline
%11 & AT-CoT $\rightarrow$ AT-SOAP & 0.4894 & 0.3403 & 0.2300 \\
%\hline
%12 & AT-SOAP & 0.4924 & 0.3430 & 0.2307 \\
%\hline
%13 & Merged Submission & \textbf{0.4986} & \textbf{0.3537} & \textbf{0.2417} \\
%\hline
%\end{tabular}
%\label{tab:final_systems}
%\end{center}
%\end{table}
%\begin{table}[t]
%\centering
%\scriptsize
%\setlength{\tabcolsep}{4pt}
%\renewcommand{\arraystretch}{1.05}
%\begin{tabular}{clccc}
%\toprule
%\textbf{\#} & \textbf{System} & \textbf{C-F1} & \textbf{R-2} & \textbf{R-3} \\
%\midrule
%10 & A-ASR $\rightarrow$ AT-SOAP   & 0.4918 & 0.3345 & 0.2256 \\
%11 & AT-CoT $\rightarrow$ AT-SOAP  & 0.4894 & 0.3403 & 0.2300 \\
%12 & AT-SOAP                       & 0.4924 & 0.3430 & 0.2307 \\
%\midrule
%13 & Merged Submission             & \textbf{0.4986} & \textbf{0.3537} & \textbf{0.2417} \\
%\bottomrule
%\end{tabular}
%\caption{Representative final systems and the merged submission model on the development set.}
%\label{tab:final_systems}
%\end{table}

\section{Discussion}

\paragraph{Official Evaluation}
The official evaluation, shown in Table~\ref{tab:final_systems}, comprises three test sets: the in-domain DoPaCo test set, the Mock Dialogue dataset \cite{Fareez2022}, and the Realistic dialogue recordings collected for the shared task. Although both submissions achieved similar development-set performance, our primary merged model, using the checkpoint averaging strategy of \cite{ugan2025weight}, consistently outperformed the contrastive submission on all official test sets, with the largest gains under increasing domain shift. These results indicate improved robustness and reduced overfitting to synthetic TTS data.

\paragraph{SOAP Data Generation Prompt Variants}
\label{subsubsec:soap_generation}
We further analyze simpler prompt variants on the Synth-DoPaCo development set, including a SOAP template prompt and a few-shot prompt with two SOAP examples. As shown in Table~\ref{tab:soap_generation_prompt}, the few-shot SOAP examples prompt obtains the best development scores, suggesting that in-domain examples are particularly useful for direct SOAP generation. Since the downstream training experiments in this work are based on the SOAP template + concept statistics prompt, we leave incorporating few-shot-generated SOAP supervision into the full training pipeline for future work.
 % examples sampled from the training data

\vspace{-3pt}
\begin{table}[htbp]
\caption{Development-set comparison of SOAP generation prompt variants using Qwen-3.5-27B.}
\begin{center}
\scriptsize
\begin{tabular}{|l|c|c|c|}
\hline
\textbf{Prompt variant} & \textbf{Concept-F1} & \textbf{R-2} & \textbf{R-3} \\
\hline
SOAP template + concept statistics & 0.4191 & 0.2481 & 0.1481 \\
\hline
SOAP template & 0.4409 & 0.2616 & 0.1562 \\
\hline
Few-shot SOAP examples & \textbf{0.4791} & \textbf{0.2916} & \textbf{0.1751} \\
\hline
\end{tabular}
\label{tab:soap_generation_prompt}
\end{center}
\end{table}

\section{Conclusion}
We presented KIT's submission to the BeTraC 2026 Lightweight Track for end-to-end speech-to-SOAP generation. Our main contribution is a scalable data augmentation pipeline that unifies heterogeneous medical dialogue datasets through synthetic speech generation and automatically generated SOAP supervision, enabling effective adaptation of Qwen2.5-Omni.

Our final merged system achieved the best performance among our submitted systems across the official test sets, indicating that averaging diversely trained checkpoints can improve robustness under domain shift.
%Our final merged system achieved the best performance among our submitted systems across the official test sets, suggesting that checkpoint averaging improves robustness under domain shift. 
% Future work will investigate stronger SOAP supervision generation, including few-shot-generated SOAP notes, and further improve generalization to realistic clinical dialogues.

% We presented KIT's submission to the BETRAC 2026 Lightweight Track for end-to-end speech-to-SOAP generation. Our main contribution is a scalable data augmentation pipeline that unifies heterogeneous medical dialogue datasets through synthetic speech generation and automatically generated SOAP supervision, enabling effective adaptation of Qwen2.5-Omni.

% The official evaluation Table~\ref{tab:final_systems} comprises three test sets: the in-domain DoPaCo test set, the Mock Dialogue dataset \cite{Fareez2022}, and the Realistic dialogue recordings collected for the shared task. Although both submissions achieved similar development-set performance, our primary merged model, using the checkpoint averaging strategy of \cite{ugan2025weight}, consistently outperformed the contrastive submission on all official test sets, with the largest gains under increasing domain shift. These results indicate improved robustness and reduced overfitting to synthetic TTS data.

\section*{Acknowledgment}
Generative AI tools were used for grammar correction, readability improvements, and \LaTeX{} formatting and structuring. All scientific content was produced and verified by the authors.
This work was supported by the project “How is
AI Changing Science? Research in the Era of
Learning Algorithms” (HiAICS), funded by the
Volkswagen Foundation, and partially by the European Union's Horizon research and innovation
programme under grant agreement No. 101135798,
project Meetween (My Personal AI Mediator for
Virtual MEETtings BetWEEN People) and European Union's Horizon Europe programme grant
agreement No. 101213369 (DVPS). The authors
gratefully acknowledge computing time provided
on HoreKa at the National High-Performance Computing Center at KIT (NHR@KIT), supported by
the Federal Ministry of Education and Research,
the Ministry of Science, Research and the Arts of
Baden-Württemberg, and the DFG.

\end{document}